\documentclass[lettersize, journal]{IEEEtran}
\usepackage{amsmath,amsfonts}
\usepackage{algorithmic}
\usepackage{array}
\usepackage[caption=false,font=normalsize,labelfont=sf,textfont=sf]{subfig}
\usepackage{textcomp}
\usepackage{stfloats}
\usepackage{url}
\usepackage{verbatim}
\usepackage{graphicx}
\usepackage{booktabs}

\def\BibTeX{{\rm B\kern-.05em{\sc i\kern-.025em b}\kern-.08em
    T\kern-.1667em\lower.7ex\hbox{E}\kern-.125emX}}
\usepackage{balance}

\begin{document}

\title{AeCoM: Design, Modeling and Control of a Novel Aerial Continuum Manipulator}

\author{Rui Peng, Student Member, IEEE, Zehao Wang and Peng Lu, Member, IEEE 
\thanks{Manuscript created on October, 2021. This work is supported by ... project. (Corresponding author: Peng Lu, email: lupeng@hku.hk) 

Rui Peng, Zehao Wang and Peng Lu are with the Department of Mechanical Engineering at The University of Hong Kong, Hong Kong, China. }}

% \mark{IEEE/ASME TRANSACTIONS ON MECHATRONICS }% ~Vol.~18, No.~9, September~2020
% \markboth{IEEE/ASME TRANSACTIONS ON MECHATRONICS }% ~Vol.~18, No.~9, September~2020
% {How to Use the IEEEtran \LaTeX \ Templates}

\maketitle

\begin{abstract}
Aerial robotic systems has raised emerging interests among researchers.
In this paper, a novel aerial manipulation system: a flying continuum robotic manipulator (AeCoM) is first proposed, to the best of authors' knowledge.
In the perspective of design, a lightweight tendon-driven continuum robotic arm (in 3D-printed material) is delicately coupled with a quadrotor.
To address the problem of kinematics inaccuracy due to different tip loading, we introduce an attitude sensor (IMU) to assist in PCC (Piecewise Constant Curvature) configuration.
To deal with frequent and complex aerial manipulation tasks, we deploy a tension-based closed-loop control method, which is used to avoid tendon-slacking in manipulating the shape of the continuum arm.
Distinct from the conventional aerial rigid manipulators, the proposed system achieve more relative payload capability and motion dexterity.
The system's experimental results validate the performance of tendon-slacking avoidance, kinematics accuracy with different tip loading, and tip positioning accuracy for aerial grasping.
The comparison with conventional aerial manipulators, indicates that the proposed manipulator has better manipulation performance and more potential applications in the cluttered environment.  
\end{abstract}

\begin{IEEEkeywords}
Aerial system, tendon-driven continuum robot, design and control, kinematics.
\end{IEEEkeywords}

\section{Introduction}

Aerial manipulation \cite{ruggiero2018aerial} has been a recent research hotpot in the robotics field, which integrates lightweight robotic manipulators with versatile aerial platforms, such as autonomous multirotors or helicopters.
Several prototypes of aerial manipulation robots \cite{kim2018origami, doyle2011avian} are designed and validated both in indoors and outdoors.
Many research works have investigated different applications like physical interaction with environment \cite{jimenez2013control, korpela2014towards, kim2015operating, tsukagoshi2015aerial, lee2020aerial, suarez2017anthropomorphic}, 
aerial grasping \cite{chen2019aerial, kim2013aerial, mellinger2011design, thomas2014toward}, 
inspection and maintenance \cite{jimenez2015aerial, tognon2019truly}, etc.
In order to enhance robustness and stability, and broaden functionalities of the aerial manipulation systems, advanced techniques like visual servoing control \cite{ruggiero2015multilayer}, motion planning and trajectory tracking control \cite{heredia2014control}, or interaction contact force control \cite{suarez2018physical}.
However, most of current aerial manipulators involve several degrees of freedom (DOFs) with rigid links, which have limits of payload capacities, uncertainty handling and motion dexterity.
For instance, it is still difficult for the existing aerial manipulation systems to execute aerial operations or interaction with objects in the constrained and cluttered environment.
Therefore, existing problems of this robotics area trigger possibility of the emerging of a generation of new aerial manipulators.

% \begin{figure}[t]
% 	\centering
% 	\includegraphics[width=\columnwidth]
% 	{1_frontfig.png}
% 	\caption{The proposed aerial continuum manipulator (AeCoM) is conducting aerial grasping operations.}
% 	\label{fig:1_frontfig}
% \end{figure}

Continuum robots have been developed rapidly in recent years, and become a rich and diverse research field with plenty of designs and demonstrated applications.
A continuum robot is usually defined as a continuously bending structure with infinite-degree-of-freedom.
Considering extinct flexibility, safety and lightweight property, it is possible to replace conventional rigid manipulators with continuum robot arms in aerial manipulation systems \cite{samadikhoshkho2020modeling}.
Inspired by aforementioned features of the continuum robots, we first propose an aerial continuum manipulation system (AeCoM) shown in Fig. \ref{fig:1_frontfig}, with original mechanical design, comprehensive forward and inverse kinematic model, and a continuum pose control method with shape manipulating.
AeCoM is composed of a quadrotor, a tendon-driven continuum robotic arm with a gripper on the end-effector, and an extra landing system.
The mechanical connection area between the quadrotor and the arm is highly compact for optimized structural layout.
The integrated system can perform autonomous flights, aerial grasping tasks and other potential operations.

Despite of inherent advantages of continuum robots, tendon-driven continuum robots can provide more precise kinematics model and higher payload capacity than other types of continuum robots \cite{yeshmukhametov2020study}.
However, tendon-driven continuum robots critically suffer tendon slacking during bending motion, even for static mechanical bases which are most of existing continuum robots based on.
Such tendon loosing harmfully affects the accuracy of kinematics model, and the motion continuity. 
Typically for AeCoM, whose mechanical base is floating with the quadrotor platform during complex aerial manipulation tasks, the risk of tendon slackening is increased dramatically, which harms the feasibility of the aerial system. 

To address the issue of tendon slacking, researchers have investigated many approaches, which could be categorized as three aspects: mechanical design, sensor assistance and numerical models.
A passive pretension mechanism is developed to solve wire slacking and derailing problems, and improve continuum robot payload capacity \cite{yeshmukhametov2020wire}.
A novel passive brake mechanism, the capstan brake, is presented to improve the energy efficiency of a tendon-driven system and prevent the tendon from escaping the spooler through the use of rollers \cite{in2012capstan}.
Stiffness adjustable tendon (SAT), a nonlinear spring with hysteresis characteristics is used to enhance tension controllers to improve safety and versatility of continuum manipulators in human environment \cite{haiya2010tension}.
A slack enabling mechanism is proposed to increase the efficiency and guarantee the safety of the soft tendon routing system, with maintenance of the tendon tension \cite{in2015feasibility}.
A hyper-redundant Pulleyless Rolling joint with Elastic Fixtures (PREF joint) is proposed to aid the actuation cables to maintain the tension \cite{suh2015design, suh2014underactuated}.
However, mechanical design based solutions bring extra weight and system complexity, and lack generalization for other continuum robot platforms. 

For sensor assistance in control of continuum robots, a sensor-driver integrated muscle module by integrating necessarily components for tendon-driven robots, is proposed to improve high-tension measurability and flexible tension control \cite{asano2015sensor}.
A closed-loop wire tension control system for a wire-EDM machine is presented to improve the accuracy of the kinematics model \cite{yan2004accuracy}.
To improve the maneuverability and hence the accuracy of the catheter tip navigation, a model-free catheter tip position control based on the position and tension feedback with a new robotic catheter system is proposed \cite{back2017model}.
A wire tension control system is designed to reduce the wire tension fluctuation, with a mathematical model of the control system based on DC servo motors, a linear motion platform, and tension sensors \cite{li2016study}.
In terms of numerical models, Kaitlin et al. analytically extend a previous nonlinear Cosserat-rod-based model for tendon-driven robots to handle prescribed tendon displacements, tendon stretch, pretension, and slack \cite{oliver2019continuum}.
Based on solid mechanics by analyzing the effects of geometrically nonlinear tendon loads, a new linear model is presented to form a concise mapping from beam configuration–space parameters to n redundant tendon displacements via the internal loads and strains, to achieve accurate control and avoid slacking tendons for manipulators \cite{camarillo2008mechanics}.
A novel analytical modeling approach is presented to investigate the combined effects of external loads (i.e., contact and gravitational loads) and internal distributed friction forces on tension loss of a generic tendon-driven continuum manipulator, to realize accurate shape or position control \cite{liu2021effect}.

Despite previous efforts in solutions of avoiding tendon slacking, there is still a lack of verifying the situations of fast and continuous continuum motion, which brings challenge for precise pose control.
In light of the observations, we introduce an inertia measurement unit (IMU) installed on the end-effector tip which provide accurate attitude information, and several torque sensors attached to tendons which return tension information of each tendon.
Based on these sensor feedback, a closed-loop control method for the aerial continuum manipulator is established to achieve precise shape control, and avoid tendon slacking during aggressive bending motion in aerial flights.
Experiments prove that there is no tendon slacking occurring in any AeCoM motions, and then it is fundamental to build a comprehensive kinematics model to compute the realtime pose of end-effector for further applications. 
Since the AeCoM is a novel aerial robotic system, its kinematics model should be based on conventional aerial manipulators \cite{meng2020survey, ruggiero2018aerial}, and involve the kinematics model of the continuum robot, which has been widely investigated \cite{webster2010design, burgner2015continuum}. 

With respect to the latter, a new method for synthesizing kinematic relationships for a general class of continuous backbone, or continuum, robots is introduced \cite{jones2006kinematics}.
A kinematic model for a wire-driven continuum manipulator without assuming constancy in curvatures of the manipulator’s shape, is proposed to completely attain the tip orientation via pulling-wire lengths regardless of its whole shape \cite{hsiao2017wire}.
Ahmad et al. propose a novel pose estimation and obstacle avoidance approach for tendon-driven multi-segment continuum manipulators moving in dynamic environments \cite{ataka2016real}.
A new variable curvature continuum kinematics model is presented for multi-section continuum robots with arbitrarily shaped backbone curves assembled from sections with three degrees of freedom (DOFs) \cite{mahl2014variable}.
However, previous studies on kinematics model only work for continuum robots without considering internal or external loading, which might cause unknown shape deformation and consequent tendon slacking. 
The primary consequence is that the end-effector pose calculation lack efficiency and accuracy, especially for different loading during aerial manipulation tasks.
Thus, it is necessary to propose a kinematics model for the AeCoM with taking loading into account.
An analytical loading model accounting for tendons and the tendon loading distribution, is derived to prevent slack in tendons for a given configuration \cite{dalvand2018analytical}.
Federico et al. present a method for improving the accuracy of kinematic models of continuum manipulators through the incorporation of orientation sensor feedback \cite{campisano2020online}.
A data-driven approach to identifying soft manipulator models that enables consistent control under variable loading conditions, is present to achieve autonomous control of a pneumatically actuated soft continuum manipulator \cite{bruder2021koopman}.
Despite of acceptable performance of these works, there still exists much online computation.
To reach a lightweight and precise kinematics model dealing with random extra loading for the AeCoM, we incorporate an IMU that gives tip orientation information, into the geometrical mapping relationship based on assumption of PCC between tendon lengths and the shape of continuum manipulator.

The main contributions of this paper can be summarized as follows:

\begin{enumerate}
	\item We first propose a novel aerial manipulation system: aerial continuum manipulator (AeCoM), with original mechanical design. The system has advantages of motion dexterity and payload capacity over conventional aerial manipulators. 
	\item With incorporation of the sensor IMU, a specific kinematics model based on PCC geometrical mapping is proposed, even under circumstances of different loading. Experiments of aerial bending motion and object grasping are conducted to prove the distinct accuracy.
 	\item By employing the IMU and torque sensors, a closed-loop control method is derived to prevent tendon slacking during aggressive bending motion.
\end{enumerate}

\section{Design}

In this section, the overview system architecture which is divided into hardware design and mechanical design, is illustrated in detail, and comparison with conventional aerial manipulators is presented.

% \begin{figure*}[t]
% 	\centering
% 	\includegraphics[width=\textwidth]
% 	{2_architeture.pdf}
% 	\caption{ Detailed mechanical system architecture. (a) (b)   . }
% 	\label{fig:2_architeture}
% \end{figure*}

\subsection{Hardware Design}

The hardware system is designed in order to build a solid basis of a complete and integrated control system.
We separate the whole aerial system into three subsystems: automatic decision system, aerial flight platform and continuum robotic system.
The first system is mainly an onboard high-performance PC, which is responsible for receiving and processing sensing information, and generating every motion commands.
The aerial platform is a quadrotor using an open-source flight controller board and motors with high payload capability.
The last one involves an independent micro-controller board as the core, tension sensors, an IMU, and several functional servo motors.
The latter two subsystems directly connect to the onboard PC via serial ports, which form stable communication systems.  
All of the key hardware components are listed in Table. \ref{tab:hardware_list}.
The workflow of the whole hardware system is depicted in Fig. \ref{fig:3_workflow1}.

\begin{table}[h]
	\begin{center}
		\caption{Key hardware components list }  \label{tab:hardware_list}

		\begin{tabular}{lcccc}
		\toprule

			Hardware components & Quantity & Model \\  
			\midrule
			Onboard PC 				& 1 & Dji manifold-v2 \\  
			Flight controller 		& 1 & PixRacer-micro \\  
			Propeller motors 		& 4 & T-motor F90 \\  
			Micro-controller board  & 1 & Stm32F1VCT6 \\  
			Tendon motors           & 4 & FeeTech-STS3032 \\
			USB-TTL communication board    & 1 & FeeTech-URT1 \\
			IMU                     & 1 & MPU9250 \\
			Digital servo motors for landing    & 2 & RDS3115 \\
			Servo motor of the gripper    & 1 & WeeTech-HWZ020 \\

		\bottomrule

		\end{tabular} 
	\end{center}
\end{table}

% \begin{figure*}[t]
% 	\centering
% 	\includegraphics[width=\textwidth]
% 	{3_workflow1.pdf}
% 	\caption{ Hardware workflow. }
% 	\label{fig:3_workflow1}
% \end{figure*}

% \begin{figure*}[t]
% 	\centering
% 	\includegraphics[width=\textwidth]
% 	{3_workflow1.pdf}
% 	\caption{ Hardware workflow. }
% 	\label{fig:3_workflow1}
% \end{figure*}

% \begin{figure*}[t]
% 	\centering
% 	\includegraphics[width=\textwidth]
% 	{4_forceDistribution.pdf}
% 	\caption{ Force distribution. }
% 	\label{fig:4_forceDistribution}
% \end{figure*}

\subsection{Mechanical Design}

To incorporate a continuum robotic system into an aerial platform, the entire weight, payload capacity and control accuracy or dexterity should be considered.
In order to reduce the weight as light as possible, soft material is first taken into account. 
Despite of the property of light weight, soft material has the deadly drawback that its stiffness is not large enough, such that the whole soft manipulator can not bear heavier objects.
Under this situation, its payload capacity is severely limited, and also the soft material is mostly driven by pneumatic pumps, which are too heavy for the aerial system.
Therefore, instead of pneumatic driving, we deploy a tendon-driven system, which is actuated by several motors.
Moreover, we use 3D-printing material to produce main mechanical components of the continuum robotic arm.

To achieve larger workspace, the prototype consists of five same segments consecutively to extend its bending range, which form one single continuum section, as shown in Fig. \ref{fig:2_architeture}.
Each segment is composed of two cross disks, which are connected by a mechanical gimbal, and four springs evenly distributed in each disk as supportance structures.
The bending motion is actuated by four tendons which are driven by four motors with torque feedback.
Of note, the key factor of production of the manipulator, is the usage of the specific motor whose weight is only $20g$.
In addition, each motor could provide the maximum torque of $3.5N\cdot m$, and its volume is $0.15 cm^3$.
In other words, due to extinct mechanical properties, the motors facilitate optimization of the system structural layout where the actuation part of the robotic arm is compacted into a highly limited space, and also supply with necessary actuation for satisfying different complicated motion.

Regarding to the space distribution of four actuation motors or tendons, one pair of diagonal tendons manipulate the robotic arm's bending motion within the plane of the tendons, while the other diagonal tendons control the perpendicular bending motion.
From the perspective of mechanical design, the gimbal only restricts two DOFs of rotation motion, such that there is no twist moment ocurred during the manipulator's motion.
Therefore, the bending motions triggered by the two pairs of diagonal tendons which are strong fish cables, are totally independent.
Considering accurate end-effector (EE) pose estimation, we employ an IMU sensor which is installed centrally on the EE tip, to obtain realtime and precise attitude information, as shown in Fig. \ref{fig:2_architeture}.
The roll axis is corresponding to one of the two pairs, and the pitch axis aligns to another, which causes that the bending motion can be detected by IMU's attitude changes.
Discussing the structural layout, every component is evenly concentrated around the center line of the whole system, which leads to the best weight balance.

Back to the vision of the whole continuum robotic arm, its total weight is less than $50g$, which is much more lightweight than most of previous aerial manipulators.
Despite of the advantage of light weight, its broken strength is still required to validate.
Thus, we build a simulation environment in the ANSYS software, where the manipulator is conducted stress tests under the static status. 
The resistance force distribution of every mechanical element is shown in Fig. \ref{fig:4_forceDistribution}.
The simulation results demonstrate that no matter what kind of motions, the arm's structure could bear the maximum force of $80N$, which is much larger than most of targeted objects.

Compared with previous continuum robots with similar structures, the proposed design uses optimized cross-sectional disks instead of solid circular disks.
The aim is to reduce unnecessary weight of the disks, and maintain enough structural strengths.
Table. \ref{tab:mechanics_list} presents information of each detailed mechanical element, including quantity, weight, material.

\begin{table}[h]
	\begin{center}
		\caption{Continuum robotic arm's mechanical components list }  \label{tab:mechanics_list}

		\begin{tabular}{lcccc}
		\toprule
			Mechanics components & Quantity & Weight & Material  \\  
			\midrule
			Starting end disk 		& 1 & $5g$   & 3D-printing \\  
			Intermediate disk 		& 4 & $3g$   & 3D-printing \\    
			Terminating end disk	& 1 & $4g$   & 3D-printing \\    
			Mechanical gimbal  	    & 5 & $2g$   & 3D-printing \\    
			Connecting shaft        & 10 & $1g$  & Metal \\  
			Supporting spring       & 20 & $0.5g$  & Metal \\  
			Actuation motor		    & 4 & $20g$   & Mixed \\  
		\bottomrule

		\end{tabular} 
	\end{center}
\end{table}

\subsection{Auxiliary design}

For most of feasible aerial manipulation systems, one structural part which can not be overlooked is the auxiliary landing rod.
The landing rods are used to support the whole aerial system, before or after conducting aerial flights, to prevent unnecessary physical damages on the manipulator.
However, static landing rods will definitely shrink the workspace of the robotic arm.
Therefore, we design a pair of landing rods which can change their positions by controlling corresponding servo motors, as shown in Fig. \ref{}.
When the aerial system is on the ground, the landing rods switch to standing mode to sustain the manipulator.
While the system is during aerial flights, the landing rods is rotated by $180^o$ pointing to the above direction, such that the manipulator's motions are not affected by them.

\section{Modeling}

This section presents the forward and inverse kinematics model, and the dynamic model of the aerial robotic system.

\subsection{Kinematics Model}

Kinematics of flight manipulator studies the spatial coordinate relationship between each motion execution module, so as to describe the motion characteristics of the whole system.
Here we know the transformation relationship between each coordinate system, so what we want to solve most is the end effector, that is, the six degree of freedom spatial pose ($ [x \quad  y \quad  z \quad qw \quad qx \quad qy \quad qz]^T $) of hand grasp in this system.

$T_{End}$ is expressed as:

\begin{equation}
	T_{End} = T^W_U   T^U_{S_1}   \Gamma^n_{i=1}(T^{S_i}_{S_{i+1}})
	\label{equ:1}
\end{equation}
or:
\begin{equation}
	\begin{aligned}
		T_{End} = T^W_U \cdot T^U_{S_b} \cdot T_{\{S\}} \\
		\label{equ:1}
	\end{aligned}
\end{equation}

$T^W_U$ denotes the world inertial coordinate system to the UAV central coordinate system, obtained by vicon.
$T^U_{S_b}$ is the UAV center coordinate system to manipulator base coordinate system, it is obtained by CAD.
$T_{\{S\}}$ represents the forward kinematics of the manipulator itself is solved based on the PCC model.

Based on the augmented continuous arm model, the transformation relationship between each joint coordinate system can be obtained as follows:
\begin{equation}
	\begin{aligned}
		T^{S_i}_{S_{i+1}} =  \left[ \begin{matrix}   \cos(\theta_i) & -\sin(\theta_i) & 0 & l_i \frac{\sin(\theta_i)}{\theta_i}\\ \cos(\theta_i) & \cos(\theta_i) & 0 & l_i \frac{\sin(\theta_i)}{\theta_i}\\ 0 & 0 & 1  & 0\\ 0 & 0 & 0  & 1\end{matrix}  \right]		\label{equ:1}
	\end{aligned}
\end{equation}
When $\theta_i =  0$,

\begin{equation}
	\begin{aligned}
		T^{S_i}_{S_{i+1}} =  
		\left[
		\begin{matrix}
		  1 & 0 & 0 & l_i \\
		  0 & 1 & 0 & 0\\
		  0 & 0 & 1 & 0\\
		  0 & 0 & 0 & 1  
		 \end{matrix}
		 \right]
		\label{equ:1}
	\end{aligned}
\end{equation}

For a continuous manipulator with N segments, it can be defined as:
\begin{equation}
	\begin{aligned}
		\mathbf q= [q_1 \quad q_2 \quad \cdot \cdot \cdot \quad q_n]^T \in \mathbb{R}^{n\times4}
		\label{equ:1}
	\end{aligned}
\end{equation}
For each random single segment:
\begin{equation}
	\begin{aligned}
		\mathbf{q_s} = [q_{s,1} \quad q_{s,2} \quad q_{s,3} \quad q_{s,4}]^T \in \mathbb{R}^4
		\label{equ:1}
	\end{aligned}
\end{equation}
According to PCC model, infinite DOFs reduces to 2 DOFs of each segment.
According to the schematic diagram of the segment, we can get its configuration space, that is, "joint space", which is defined as:
\begin{equation}
	\begin{aligned}
		\mathbf{\lambda_s} = [\alpha_s \quad \beta_s]^T, \quad \alpha_s \in [0, \alpha_{max}], \beta_s \in (-\pi, \pi)		\label{equ:1}
	\end{aligned}
\end{equation}
Thus, the displacement of the end of the segment relative to the base is:
\begin{equation}
	\begin{aligned}
		\mathbf{p^{s-1}_s} = \frac{L_s}{\alpha_s}
		\left[
		\begin{matrix}
		   (1 - \cos\alpha_s)\cos\beta_s  \\
		   (1 - \cos\alpha_s)\sin\beta_s  \\
		  \sin\alpha_s
		  \end{matrix}
		  \right]
	\end{aligned}
\end{equation}
But when $\alpha_S$ equals to 0, the singularity point will appear, so in order to ensure the calculation accuracy, the sixth-order Taylor expansion is used here:
\begin{equation}
	\begin{aligned}
		\mathbf{p^{s-1}_s}  = 
		\left[
		\begin{matrix}
		  x^{s-1}_s  \\
		  y^{s-1}_s \\
		  z^{s-1}_s
		  \end{matrix}
		  \right]
		  =
		  L_s
		\left[
		\begin{matrix}
		   \frac{\alpha_s\cos\beta_s(\alpha_s^4-30\alpha_s^2+360)}{720} \\
		   \frac{\alpha_s\sin\beta_s(\alpha_s^4-30\alpha_s^2+360)}{720} \\
		  \frac{(\alpha_s^4-20\alpha_s^2+120)}{120} \\
		  \end{matrix}
		  \right]
	\end{aligned}
\end{equation}
At the mean time:
\begin{equation}
	\begin{aligned}
		\mathbf{\lambda_s} =
		\left[
		\begin{matrix}
		  \alpha_s  \\
		  \beta_s 
		  \end{matrix}
		  \right]
		  =
		  \left[
		\begin{matrix}
		  2\arccos(\frac{|z^{s-1}_s|}{||p^{s-1}_s||_2})  \\
		  \arctan(\frac{y^{s-1}_s}{x^{s-1}_s})
		  \end{matrix}
		  \right]
	\end{aligned}
\end{equation}
Here, the mutual conversion from configuration (joint space) to operation (operation space) of the manipulator is completed.
Because the manipulator is cable driven, an important relationship can be obtained according to the spatial relationship in the figure:
\begin{equation}
	\begin{aligned}
		L_{s,i} = L_s - \alpha_s r \cos[\beta_s+(i-1)\mu], \quad \mu = \frac{\pi}{2}, i = {1,2,3,4}
	\end{aligned}
\end{equation}
According to the diagram above, we can get the relationship between $L_ S $ and $l_{s, 1} $is:
\begin{equation}
	\begin{aligned}
		L_s = \alpha_s t_1, \quad L_{s,1} = \alpha_s t_0 \\
		t_1 - t_0 = r, \quad  L_s - L_{s,1} = \alpha_s r \\
		L_{s,1} = L_s - \alpha_s r \\ 
		L_{s,i} = L_s - \alpha_s r \cos[\beta_s+(i-1)\mu] \\
	\end{aligned}
\end{equation}
That is, the varying lengths of the four actuator lines of the segment are obtained through the joint space.
Thus:
\begin{equation}
	\begin{aligned}
		\mathbf{ q_{s,i} }= - \alpha_s r \cos[\beta_s+(i-1)\mu] \\
		% \left\{  
					% \begin{array}{**lr**}  
		q_{s,1}= - \alpha_s r \cos[\beta_s] \\
		q_{s,2}= - \alpha_s r \cos[\beta_s+\mu] \\
		q_{s,3}= - \alpha_s r \cos[\beta_s+2\mu] \\
		q_{s,4}= - \alpha_s r \cos[\beta_s+3\mu] \\  
					%  \end{array}  
		% \right.
	\end{aligned}
\end{equation}
That is, the line length changed after the actuator is driven, which establishes the relationship between acutation and configuration, that is, geometric mapping
Using the idea of differential element, there are:
\begin{equation}
	\begin{aligned}
		\mathbf{\dot{q}_s} \approx \Delta \mathbf{q_s}
	\end{aligned}
\end{equation}
Then the expression of instantaneous kinematics is obtained:
\begin{equation}
	\begin{aligned}
		\mathbf{\dot{q}_s} = \mathbf{J}_{sq\lambda} \mathbf{\dot{\lambda}_s} \\
	\end{aligned}
\end{equation}
The Jacobian matrix is:
\begin{equation}
	\begin{aligned}
		\mathbf{J}_{sq\lambda} =
		\left[
		\begin{matrix}
		  -r\cos(\beta_s)      &  r\alpha_s \sin(\beta_s)\\
		  -r\cos(\beta_s+\mu)  &  r\alpha_s \sin(\beta_s+\mu)\\
		  -r\cos(\beta_s+2\mu) &  r\alpha_s \sin(\beta_s+2\mu)\\
		  -r\cos(\beta_s+3\mu) &  r\alpha_s \sin(\beta_s+3\mu)\\
		\end{matrix}
		\right]
	\end{aligned}
\end{equation}
According to the conversion relationship between configuration and operation (above), the instantaneous relationship between acutation and operation can also be obtained:
\begin{equation}
	\begin{aligned}
		\mathbf{\dot{q}_s} = \mathbf{J}_{sq\lambda}\Delta \mathbf{\lambda_s}= \mathbf{J}_{sq\lambda}(\Delta \mathbf{p^{s-1}_s})
	\end{aligned}
\end{equation}
By analyzing the local segment, the kinematics of the manipulator of the whole n-segment can be obtained:
\begin{equation}
	\begin{aligned}
	\left\{  
		\begin{array}{lr} 	
		\mathbf{p}^0_s =
		\left\{  
			\begin{array}{lr} 
			p^{s-1}_s \quad \quad\quad \quad \quad \quad ,\quad s=1 \\
			p^{0}_{s-1} + R^0_{s-1} \cdot p^{s-1}_s, \quad s>1      \\
			\end{array}  
		\right.
		\\
		\mathbf{R}^0_s =
		\left\{  
			\begin{array}{lr} 
			R^{s-1}_s \quad \quad \quad ,\quad s=1   \\
			R^{0}_{s-1} \cdot R^{s-1}_s, \quad s>1   \\
			\end{array}  
		\right.
		\\
		\mathbf{H}^0_s =
		\left[
			\begin{matrix}
			\mathbf{R^0_s}  &  \mathbf{p^0_s} \\
			\mathbf{0_{1\times3}}  &  1      \\
			\end{matrix}
		\right]
		\end{array}  
	\right.
	\end{aligned}
\end{equation}

\subsection{Dynamics Model}

Firstly, the configuration of the whole system needs to be defined, which includes the attitude of the UAV plane (Euler angle), the three-dimensional coordinates of the UAV center and the rotation angle of the manipulator, including:
\begin{equation}
	\begin{aligned}
		p^T &= [x \quad y \quad z]^T
		\\	
	  \Phi^T &= [\varphi \quad \theta \quad \psi]^T 
		\\	
	  \delta^T &= [\delta_1 \quad \delta_2 \quad \delta_3 \quad \delta_4 \quad \delta_5]^T	
	\end{aligned}
\end{equation}
Summarized as a vector, expressed as:
\begin{equation}
	\begin{aligned}
		q^T = [p^T \quad \Phi^T \quad \delta^T]^T
	\end{aligned}
\end{equation}
Then, the stress of the system is analyzed from the perspective of energy and work. Here, the Lagrange dalambert equation is used: lagrange-d'alembert, whose unified form is:
\begin{equation}
	\begin{aligned}
		\frac{d}{dt}\frac{\partial \zeta}{\partial \dot{q}}-\frac{\partial \zeta}{\partial q} = \tau + \tau_{ext}
	\end{aligned}
\end{equation}
Here, $\alpha $is defined as the kinetic energy of the system, and $\ Nu $is defined as the potential energy of the system
\begin{equation}
	\begin{aligned}
		\alpha &= \alpha_U + \Sigma^5_{i=1} \alpha_{S,i} \\
		\beta  &= \beta_U  + \Sigma^5_{i=1} \beta_{S,i} \\
	\end{aligned}
\end{equation}
or:
\begin{equation}
	\begin{aligned}
		\alpha_U &= \frac{1}{2}\dot{p}^Tm_U\dot{p} + \frac{1}{2}\dot{\Phi}^TT^TR_tI_bR^T_tT\dot{\Phi} \\
		\alpha_{S,i} &= \frac{1}{2}\dot{p}_{i}^Tm_{i}\dot{p}_{i} + \frac{1}{2}\dot{\omega_i}^T (R_TR_i) I_{i} (R_TR_i)^T \dot{\omega_i} \\
	\end{aligned}
\end{equation}
Potential energy side:
\begin{equation}
	\begin{aligned}
		\beta = m_Ugp + \Sigma^5_{i=1}m_ig(p + R_tp^U_i)	
	\end{aligned}
\end{equation}
Considering that the flight manipulator is a continuous robotic arm of cable driven, and the spring is introduced as the structural support, the force of the spring on each joint needs to be considered. Here:
\begin{equation}
	\begin{aligned}
		F_{S,i} = 4 k \cdot \delta_i \cdot \Delta_i
	\end{aligned}
\end{equation}
Finally:
\begin{equation}
	\begin{aligned}
		M(q)\ddot{q} + C(q,\dot{q})\dot{q}+G(q)+S(q) = \tau + \tau_{ext}
	\end{aligned}
\end{equation}
$\tau_ {ext} $is the external disturbance, captured objects and other additional disturbances.

\section{Control}

This section presents the control framework of the proposed system.
To address the control problem of the system, the most important application is to realize accurate positioning control.
To achieve this, we separate the whole system into two control subsystems: the aerial platform control and the continuum manipulator control.
Also, the controller of the continuum part is designed to maintain tendon tension for slacking avoidance, during any aerial manipulation tasks.

\subsection{UAV Control}

For the UAV control, a nonlinear robust adaptive hierarchical sliding mode control approach is used due to its robustness to inertial parameter uncertainties. 
Parameters of interest and subject to changes are inertial parameters of the continuum robotic arm.
In this system, the employed control method consists of two loops where the outer loop is adopted for position-velocity control and the inner loop accounts for the attitude control. 
In the two loops, adaptive updates are exploited to assure bounded estimations for inertial parameter uncertainties.

To control the position-velocity of the quadrotor, a position sliding surface $S_p$, is is defined as:
\begin{equation}
	\begin{aligned}
		S_p = v_e + K_{pos} \cdot p_e,
	\end{aligned}
\end{equation}

where $p_e$ and $v_e$ are quadrotor position and velocity errors.
By defining $U_{quad} = T_{quad} R_{quad} e_3$ and using the sliding surface, position control signal can be defined as
\begin{equation}
	\begin{aligned}
		U_{quad} = \hat{m}_{quad} (ge_3 + \dot{p}_c K_{pos} v_e) - C_{pos} s_p,
	\end{aligned}
\end{equation}
where $K_{pos}$ and $C_{pos}$ are positive definite gain matrices, and $\hat{m}_{quad}$ is the mass estimation of the quadrotor which can be estimated by
\begin{equation}
	\begin{aligned}
		\hat{m}_{quad} =  \lambda_ \theta S_p
	\end{aligned}
\end{equation}

In the above equation, $\lambda m > 0$ and $ \theta = K_{pos} v_e e ge_3 3pc$.
The stability of the proposed position control can be proved by defining the Lyapunov function and its derivative as follows:
\begin{equation}
	\begin{aligned}
		V_p =\frac{1}{2} m_{quad} S^T_p S_p + \frac{1}{2\lambda m} m^2_{quad} ,	
	\end{aligned}
\end{equation}
By defining $U_{quad} = [ Ux Uy Uz ]^T$ from the position control command, desired quaternion Qc for attitude control
loop can be obtained as $Q_c = [\sigma_c, q^T_c]^T$, where
\begin{equation}
	\begin{aligned}
	\left\{  
		\begin{array}{lr} 	
			\sigma_c = \sqrt{\frac{1}{2} + \frac{U_z}{2||U_{quad}||}} \\
			q^T_c = \frac{1}{2||U_{quad}|| \sigma_c} [U_y \quad  -U_x \quad 0]^T
		\end{array}  
	\right.
	\end{aligned}
\end{equation}

It is notable that $||U_{quad}||$ is the norm of thrusts for motors of the quadrotor. 
This term, $||U_{quad}||$, only becomes zero if all motors are off and there is no active controller. 
Since the quadrotor always needs thrust to overcome its weight and be able to fly, this condition never happens. 
In order to control the quadrotor attitude, an attitude sliding surface, $S_q$, is defined as $S_q = \omega_e + K_q q_e$. 
Here we and $q_e$ are errors for angular velocity and quaternion, respectively. 
By using this sliding surface, an attitude control signal can be obtained as $\tau_{quad} = -\omega_{quad} \hat{J}_{quad} - C_q S_q$.

\subsection{Arm Control}

A novel arm control is proposed here for tracking the position control of the end effector of CR attached to UAV, subject to the free motion.
To design the end-effector position tracking controller, the dynamic equation of the end-effector should be derived. 
By defining the position of the end effector as $p_{end}$, its velocity, $V_{end}$, and acceleration, $a_end$ can be found as
\begin{equation}
	\begin{aligned}
		V_{end} = (p_t)_{end} =(R_q)_{end}, \quad a_{end} = (R_tq + R_qt)_{end}.
	\end{aligned}
\end{equation}
where $q_t$ should be calculated by recalling (3) and (9) as
\begin{equation}
	\begin{aligned}
		q_t = \frac{1}{\rho_A} (R^T (n_s + f_{ext} + f_{uav} + R^T f_{tendon}) - \omega q.
	\end{aligned}
\end{equation}

Substituting (35) into (34), and using the result in (33), the end effector acceleration can be found as
\begin{equation}
	\begin{aligned}
		a_{end} = \frac{1}{\rho_A} (n_s + f_{ext} + f_{uav} - \alpha_M T_M )_{end} = a_c + b_c U
	\end{aligned}
\end{equation}
where $a_c$, $b_c$, and $U$ are defined as follows:
\begin{equation}
	\begin{aligned}
		a_c &= \frac{1}{\rho_A}(n_s + f_{ext} + f_{uav})_{end} , b_c = \frac{1-}{\rho_A} (\alpha_M)_{end} \\
		U &= T_M .
	\end{aligned}
\end{equation}
To design a controller (Fig. 2) to satisfy the robustness and adaptivity requirements, the sliding surface S is defined as:
\begin{equation}
	\begin{aligned}
		S = \dot{p}_{end} - \dot{p}_r = \dot{e} + \alpha_e,
	\end{aligned}
\end{equation}
where$ \dot{p}_r = \dot{p} - 3_e$, $p_d$ denotes the desired position of the end-effector, and 3 is a positive definite matrix. Therefore,
the control law can be proposed as:
\begin{equation}
	\begin{aligned}
		U = \dot{b}^{-1}_c  \dot{p}_r - \dot{b}^{-1}_c \hat{a}_{c} - K_v \dot{e} - K_pe + \hat{\Delta}, \quad \dot{\hat{\Delta}} = -KS.
	\end{aligned}
\end{equation}

Here$ \hat{\Delta} $denotes the estimation of modeling uncertainties, while $\dot{b}_c$ and $\dot{a}_c$ are the nominal values of $b_c$ and $a_c$, respectively. 
Also $K_p$, $K_v$ and $K_0$ denote positive definite gain matrices. 
By substituting the suggested control law (39) into the end effector dynamics (36), the closed loop dynamics of end effector for ACMS arm can be derived as:
\begin{equation}
	\begin{aligned}
		\dot{b}^{-1}_c \dot{S} + K_v \dot{e} + K_pe = \widetilde{\Delta}
	\end{aligned}
\end{equation}
where $ \widetilde{\Delta} = \hat{\Delta} - \Delta $ represents the error on modeling and uncertainty estimation.
The uncertainty $\Delta$ is defined as
\begin{equation}
	\begin{aligned}
		\Delta = -\widetilde{b}_c^{-1} a_{end} + \widetilde{b}_c^{-1} \widetilde{a}_c.
	\end{aligned}
\end{equation}
Here $\widetilde{}$ denotes the error on modeling and uncertainty estimation. 
It is also assumed that the uncertainty parameter, $\Delta$ is slowly varying.

\section{Experiments}

This section demonstrates detailed comparison with previous aerial manipulators, control performance evaluation including control response and tendon slacking avoidance, validation of the kinematics model, and aerial grasping evaluation.

\subsection{Manipulator Comparison}

The comparison is divided into two aspects: comparison with conventional aerial manipulators, and comparison with previous continuum robotic arms.

\subsection{Control Performance Evaluation}

\begin{figure}[h]
	\centering
	\includegraphics[width=\columnwidth]
	{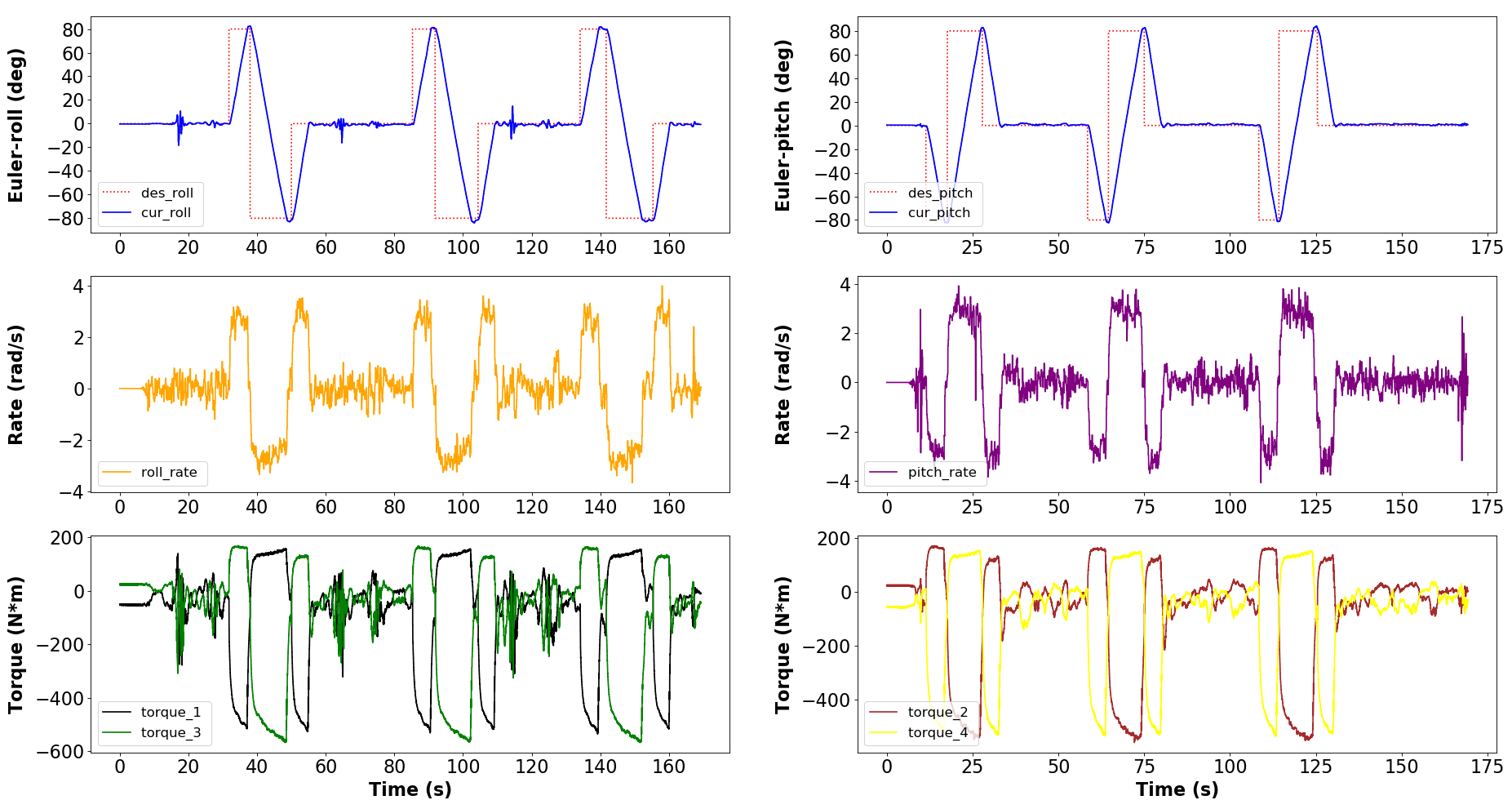}
	\caption{Aerial bending motion with predefined orientations: pitch and roll.}
	\label{fig:3}
\end{figure}

\begin{figure}[h]
	\centering
	\includegraphics[width=\columnwidth]
	{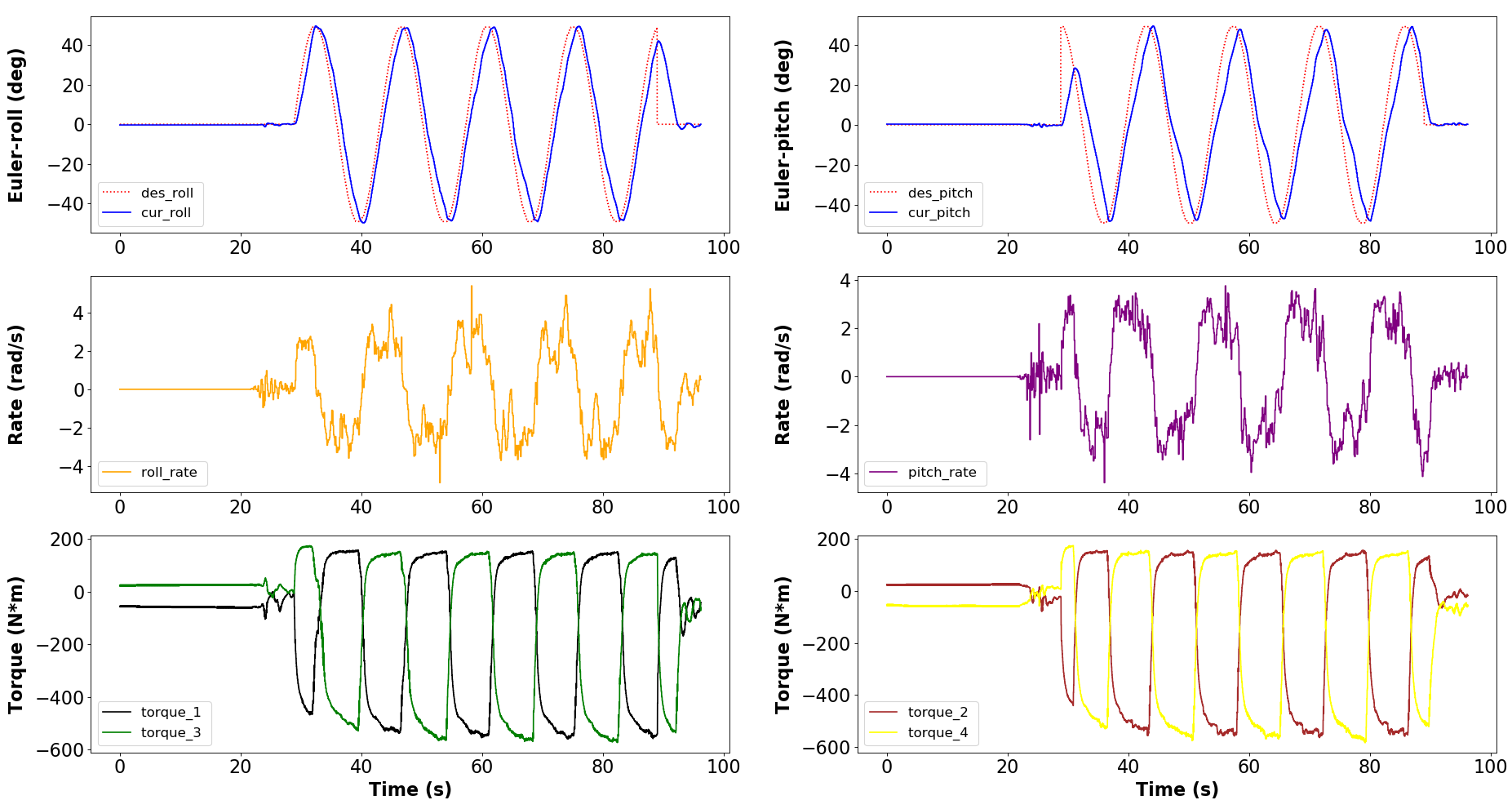}
	\caption{Aerial dynamic motion for drawing a circle.}
	\label{fig:2}
\end{figure}

\begin{figure}[h]
	\centering
	\includegraphics[width=\columnwidth]
	{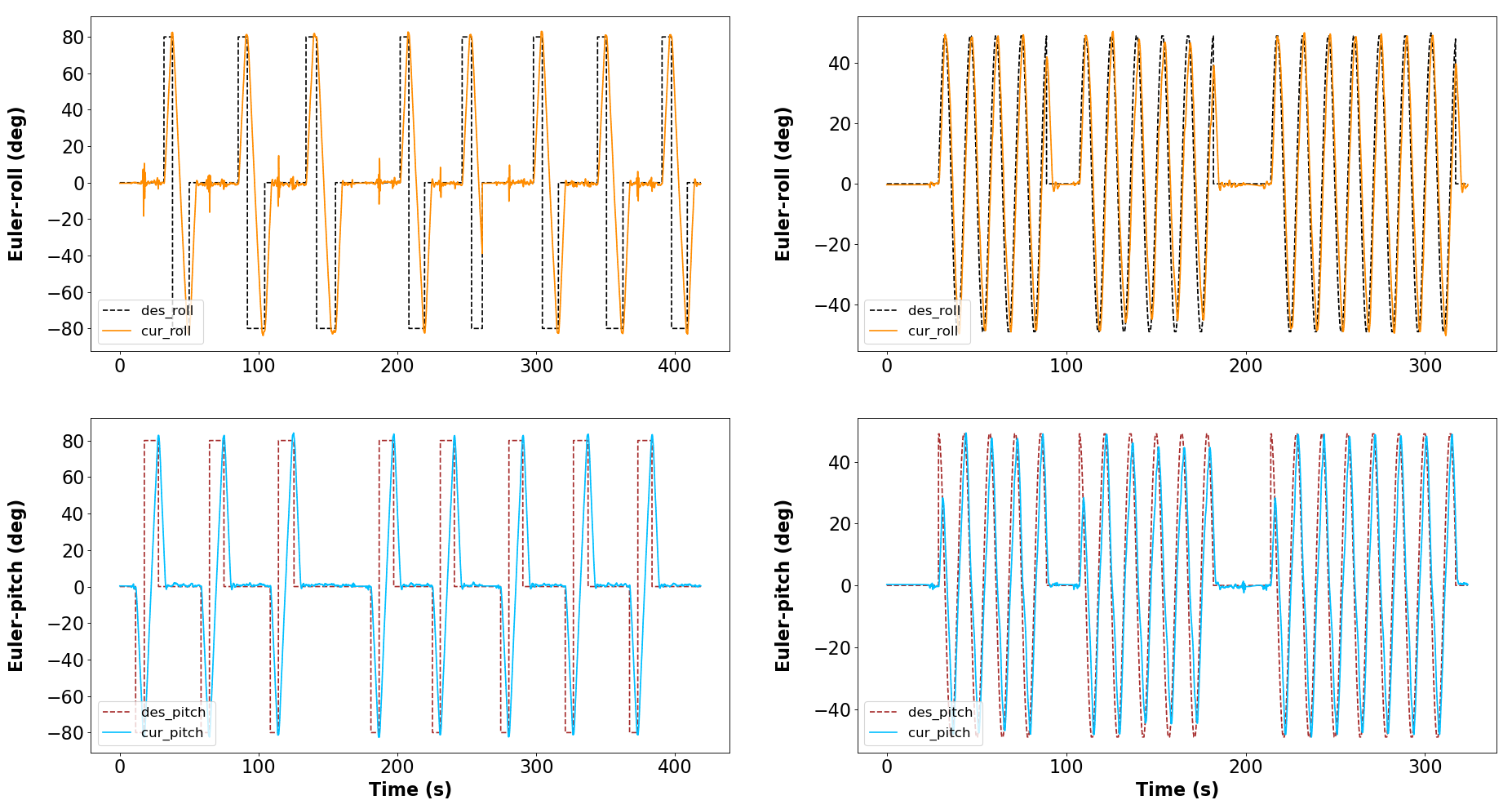}
	\caption{Repeatability of tendon slacking avoidance tests.}
	\label{fig:4}
\end{figure}

\begin{figure}[h]
	\centering
	\includegraphics[width=\columnwidth]
	{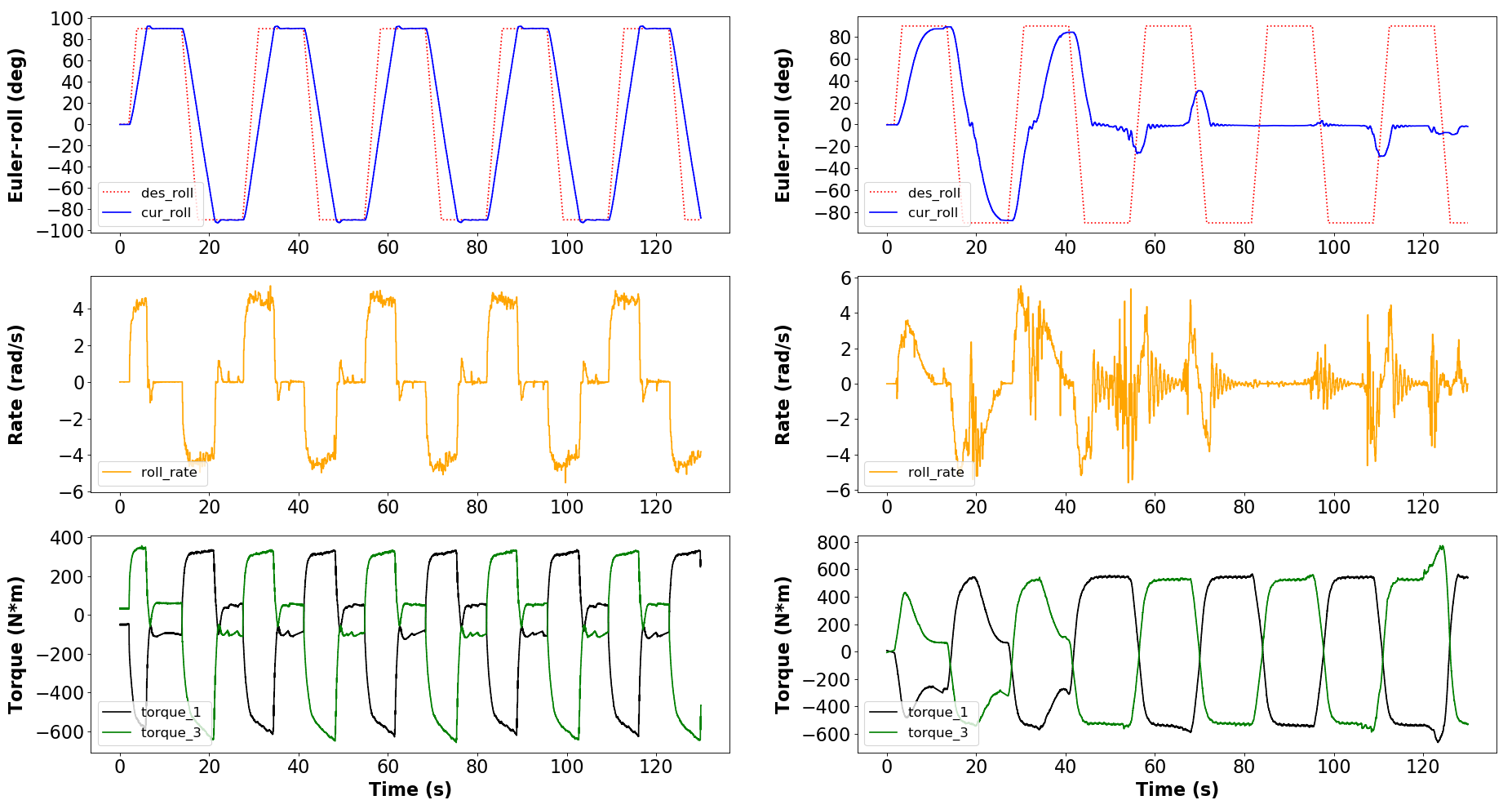}
	\caption{Comparison between involved tension feedback control and non-tension feedback control.}
	\label{fig:5}
\end{figure}

\subsection{Kinematics Model Validation}

\begin{figure}[h]
	\centering
	\includegraphics[width=\columnwidth]
	{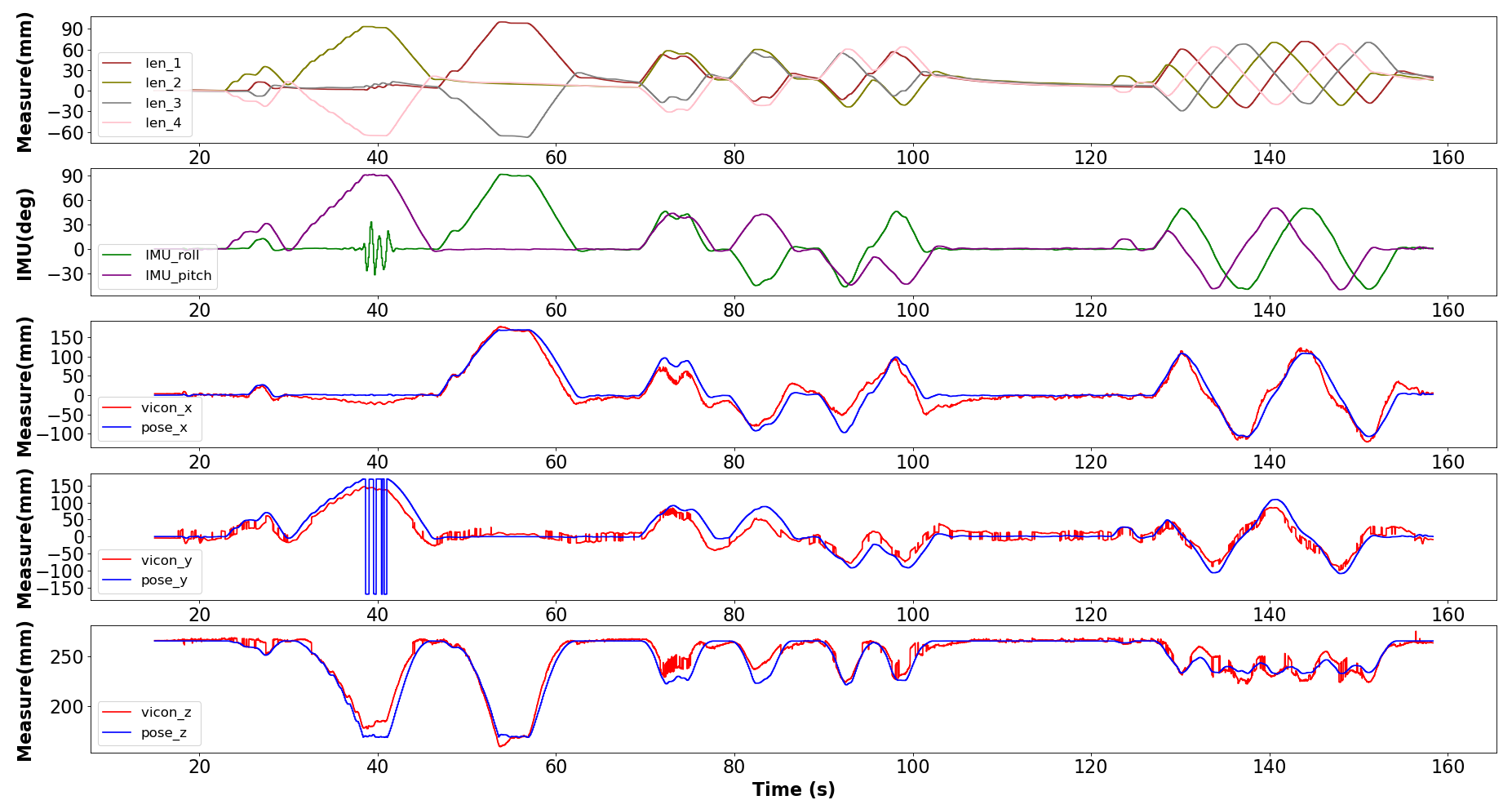}
	\caption{Forward kinematics model evaluation.}
	\label{fig:5}
\end{figure}

\subsection{Aerial Grasping}

\section{Conclusion}

This paper proposes a novel aerial continuum manipulator with original mechanical design,and complete modeling and control framework for the first time, to the best author's knowledge.
The main contributions are to address the issue of slacking for the tendon-driven continuum robotic arms, and establish a new kinematics model for the proposed aerial system.
To validate the system performance, we design several experiments: continuous aerial bending motion, aerial grasping tasks, and other comparison experiments.
The system shows good accuracy in kinematics modeling and actual applications, and potential in future works.

\balance

\bibliographystyle{IEEEtran}
\bibliography{root.bib}

\end{document}